\documentclass{article}
\usepackage{spconf,amsmath,epsfig}

\usepackage{times}
\usepackage{epsfig}
\usepackage{graphicx}
\usepackage{amsmath}
\usepackage{amssymb}
\usepackage{caption}
\usepackage{color}
\usepackage{subfig}
\usepackage{multirow}
\usepackage{pifont}
\usepackage{hyperref}
\usepackage{float}
\makeatletter
\def\UrlAlphabet{%
      \do\a\do\b\do\c\do\d\do\e\do\f\do\g\do\h\do\i\do\j%
      \do\k\do\l\do\m\do\n\do\o\do\p\do\q\do\r\do\s\do\t%
      \do\u\do\v\do\w\do\x\do\y\do\z\do\A\do\B\do\C\do\D%
      \do\E\do\F\do\G\do\H\do\I\do\J\do\K\do\L\do\M\do\N%
      \do\O\do\P\do\Q\do\R\do\S\do\T\do\U\do\V\do\W\do\X%
      \do\Y\do\Z}
\def\UrlDigits{\do\1\do\2\do\3\do\4\do\5\do\6\do\7\do\8\do\9\do\0}
\g@addto@macro{\UrlBreaks}{\UrlOrds}
\g@addto@macro{\UrlBreaks}{\UrlAlphabet}
\g@addto@macro{\UrlBreaks}{\UrlDigits}
\makeatother

\begin{document}\sloppy

\def\x{{\mathbf x}}
\def\L{{\cal L}}

\title{Difficulty-aware image super resolution via Deep Adaptive Dual-Network
}

%
\name{Jinghui Qin~\thanks{Code and results are available at: \color{blue}{\url{https://github.com/xzwlx/Difficulty-SR}}}, Ziwei Xie, Yukai Shi, Wushao Wen}
\address{School of Data and Computer Science, Sun Yat-sen University, Guangzhou 510006, China\\
\small {\{qinjingh,xiezw5,shiyk3\}@mail2.sysu.edu.cn, wenwsh@mail.sysu.edu.cn}}
\maketitle

\begin{abstract}
Recently, deep learning based single image super-resolution(SR) approaches have achieved great development. The state-of-the-art SR methods usually adopt a feed-forward pipeline to establish a non-linear mapping between low-res(LR) and high-res(HR) images. However, due to treating all image regions equally without considering the difficulty diversity, these approaches meet an upper bound for optimization. To address this issue, we propose a novel SR approach that discriminately processes each image region within an image by its difficulty. Specifically, we propose a dual-way SR network that one way is trained to focus on easy image regions and another is trained to handle hard image regions. To identify whether a region is easy or hard, we propose a novel image difficulty recognition network based on PSNR prior. Our SR approach that uses the region mask to adaptively enforce the dual-way SR network yields superior results. Extensive experiments on several standard benchmarks (e.g., Set5, Set14, BSD100, and Urban100) show that our approach achieves state-of-the-art performance.
\end{abstract}
\section{Introduction}
\label{sec:intro}

Single image super-resolution(SISR)~\cite{freeman2000learning}, has gained great research attention for decades, because it has been used in various computer vision applications, such as face hallucination~\cite{cao2017attention}, object detection~\cite{li2019cross}, video compression~\cite{lu2018dvc}, etc. As a typical ill-posed issue, Single Image Super-Resolution(SISR) aims to generate a visually clear high-resolution image $I_{SR}$ from its corresponding single low-resolution image $I_{LR}$.

Recently, deep learning based image enhancement methods~\cite{srcnn,fsrcnn,edsr,shi2016,shi2017structure,rcan,ignatov2018pirm,liu2018multi} have achieved significant improvements over conventional SR methods on restoration quality. Among these methods, Dong et al.~\cite{srcnn} proposed SRCNN, a three-layer CNN, to make the first attempt to learn a nonlinear mapping between LR and HR for image SR. To accelerate the training and testing of image SR, FSRCNN~\cite{fsrcnn} extracts features from the LR inputs and upscales spatial resolution at the tail of the network. Lim et al.~\cite{edsr} built huge SR model called EDSR by using simplified residual blocks and achieved great improvement on restoration quality. LapSRN~\cite{lapsrn}, based on a cascaded CNN framework, takes an LR image as input and reconstructs different scale SR image for restoration, progressively. Zhang et al.~\cite{rdn} proposed Residual Dense Network(RDN) based on residual dense blocks(RDB) to fully exploit the hierarchical features from all the convolutional layers. Although each image region has different difficulty, above methods process them equally, thus the representational ability of CNNs in SR task is limited. To address this problem, RCAN~\cite{rcan} proposed a residual in residual(RIR) to solve the difficulty of training deep SR network and a channel attention mechanism to improve the representation ability of SR network by discriminately treating the abundant low-frequency information across channels. As their approach can discriminately process image across channels, they still fail to settle the difficulty diversity on geometry, which has great potential to present a high-quality image SR.

\begin{figure}[t] 
	\centerline{\includegraphics[width=8.5cm]{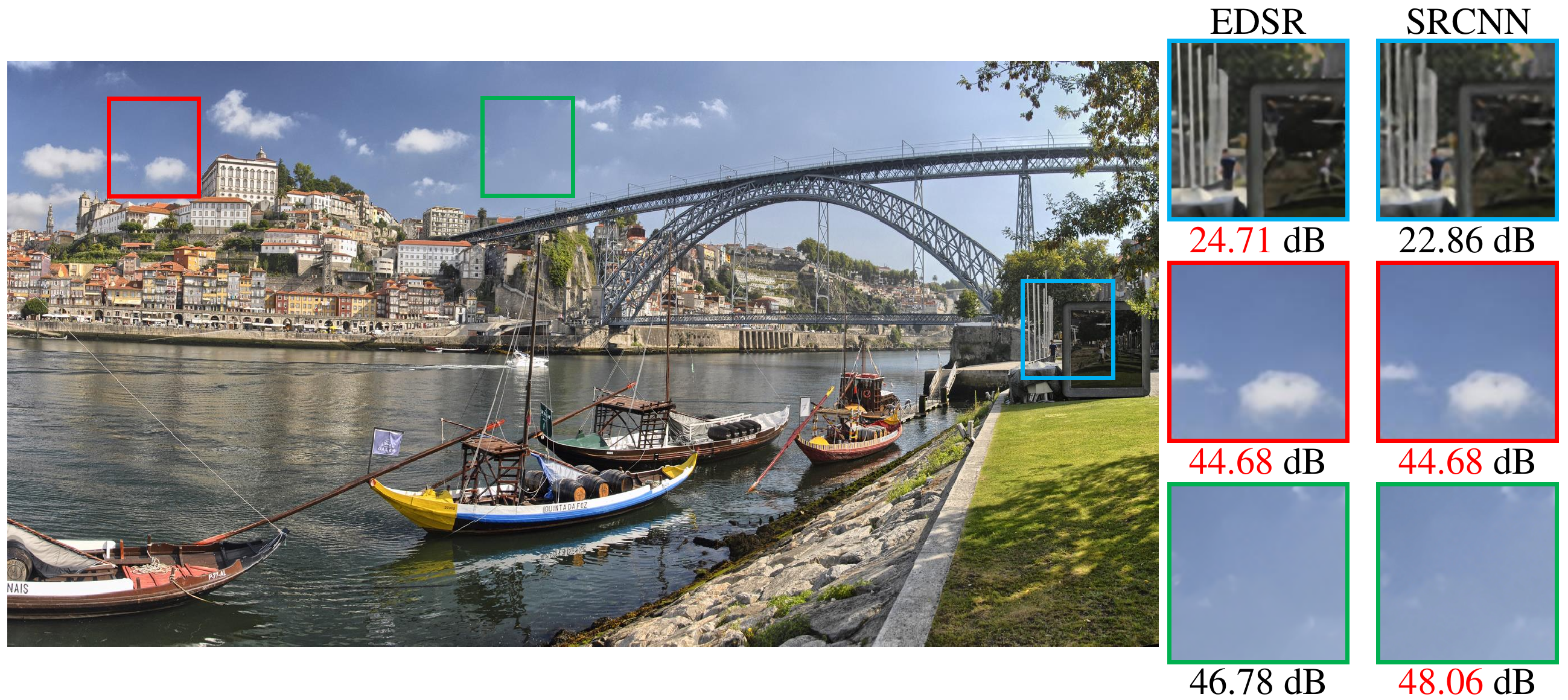}}
	\caption{Qualitative comparison with EDSR and SRCNN. EDSR reconstructs more clearly on the complex/hard region than SRCNN, but it reconstructs worse on the smooth/plain region than SRCNN. This reveals that it is suboptimal to incorporate a single model to process all regions.}
	\label{fig:example}
\end{figure}

\begin{figure*}[t] 
	\centerline{\includegraphics[width=0.99\textwidth]{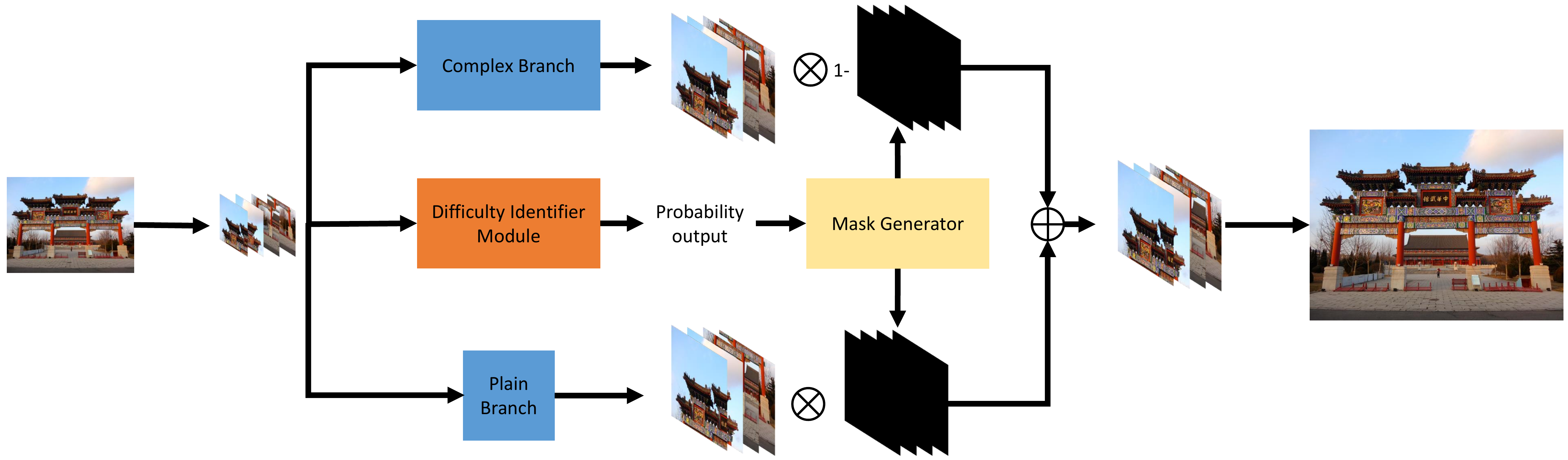}}
	\caption{Overview of our dual-way SR framework. Our framework consists of three key components: difficulty identifier module(DIM), mask generator and a dual-way SR network. Our dual-way SR network consists of two branches: a complex branch(CB) and a plain branch(PB). The CB is used to restore complex/hard patches as the PB is adopted to reconstructs plain/easy samples. DIM enforces the dual-way SR to yield superior results by applying different difficulty images into different branches.}
\label{fig:arch}
\vspace{-4mm}
\end{figure*}

Although above deep learning-based SR methods bring significant improvement of SISR, they use a unified model to process all regions of an image without considering the difficulty diversity at region-level. Generally, an image usually consists of some complex regions and some smooth/plain regions, but the difficulty of reconstructing them to high-resolution regions is not equal. As shown in Fig.~\ref{fig:example}, EDSR~\cite{edsr} presents more superior results on complex/hard regions than SRCNN~\cite{srcnn}, but it demonstrates poor restoration on smooth/plain regions. This shows that the difficulty level of all regions of an image is complicated. Therefore, it is suboptimal to use a single CNN to process all regions within an image. While a heavy model may reconstruct complex texture regions more accurately than a simple one, the simple model still shows better restoration quality in some regions. To address this problem, we propose a novel difficulty-aware region-based SR approach that uses a dual-way SR network to demonstrate a difficulty-adaptive SR process. In our dual-way SR network, one way is trained to handle easy image regions better and another is trained to handle hard image regions better. To identify the difficulty of image regions, we propose a novel image difficulty recognition network based on PSNR prior that we observed in the SR task. Our SR approach will use the region mask produced by our difficulty recognition method to adaptively enforce the dual-way enhancement SR model for accurate image SR.

In summary, the main contributions of this paper can be summarized as follows. First, we propose an image difficulty recognition network, which fully explores the PSNR prior knowledge to present a precise difficulty categorization.
Second, we propose a novel difficulty-aware SR approach that can discriminately treat each region of an image for accurate SR. With the difficulty recognition network, our dual-way SR network exhibits a high-quality restoration by alternatively utilizing different branches. Third, extensive experiments demonstrate that our approach achieves state-of-the-art performance on several standard benchmarks.

\label{sec:dimg}
\begin{figure*}[t] 
	\begin{minipage}[c]{0.48\linewidth}
  		\centerline{\includegraphics[width=0.95\textwidth]{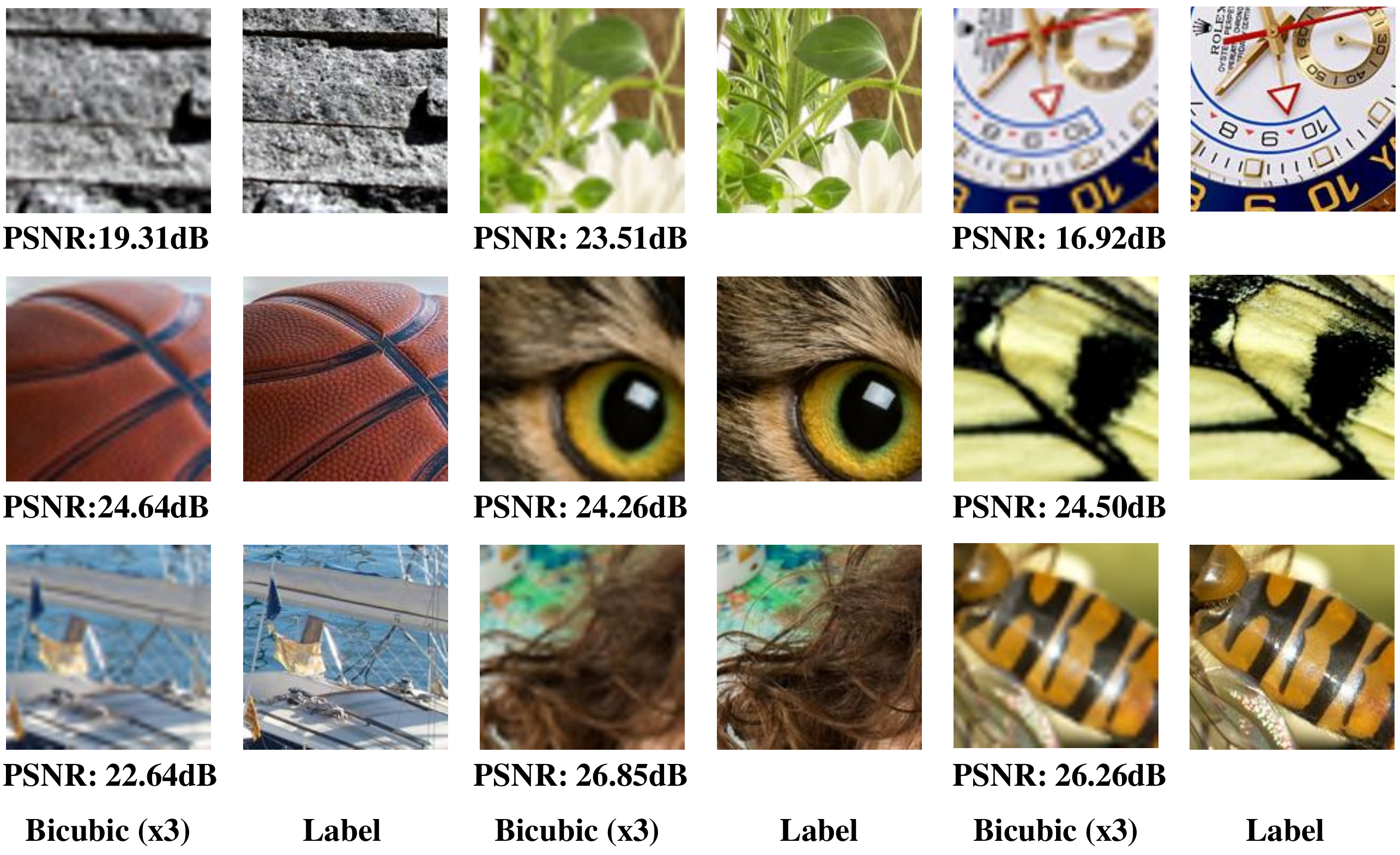}}
  		\centerline{(a) hard image patches}
	\end{minipage}
	\begin{minipage}[c]{0.48\linewidth}
  		\centerline{\includegraphics[width=0.95\textwidth]{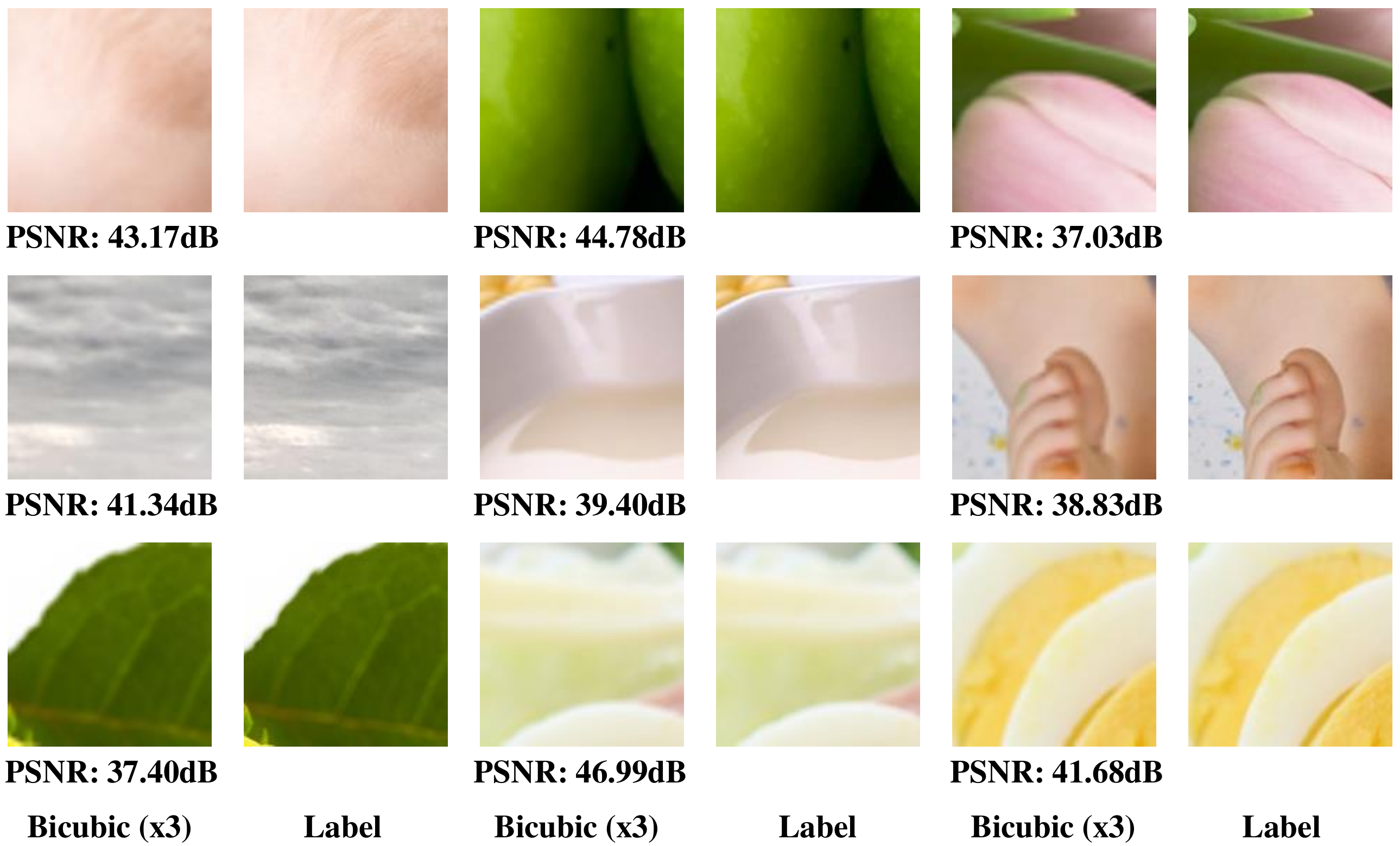}}
  		\centerline{(b) easy image patches}
	\end{minipage}
	\caption{Examples of hard and easy image patches. We cropped some images from DIV2K into patches with the size of 48$\times$48 , interpolated LR patches with Bicubic, and computed PSNR values between interpolated patches and their corresponding HR. It can be observed that hard image patches usually own low PSNR score while easy image patches tend to have high PSNR value.}
	\label{fig:example}
	\vspace{-3mm}
\end{figure*} 

\section{Methodology}
\label{sec:sisrnd}
\subsection{Framework Overview}
To treat image regions based on its difficulty discriminately, we propose a novel multi-branch SR framework that can be trained to perform accurate super-resolution. As illustrated in Fig.~\ref{fig:arch}, our proposed SR framework consists of three key components: 1) difficulty identifier module(DIM); 2) mask generator; and 3) a dual-way SR network. DIM is used for identifying the super-resolution difficulty of image regions/patches. We will detail our difficulty identifier and mask generator in Sec ~\ref{sec:dimg}. Our dual-way SR network consists of two independent SR model, complex SR branch and plain SR branch, denoted as CB and PB respectively. In our framework, CB is trained to restore hard patches while PB is dedicated to reconstructing easy patches. Unlike other SR methods that use a full-size image SR inference, our framework demonstrates a novel SR procedure as follows. First, we divide an LR full image into patches with the size of 48$\times$48. Then we input the patches into DIM. DIM generates a difficulty probability vector for each patch, where each item in the vector represent the possibility a patch belongs to the corresponding difficulty level. CB and PB reconstruct HR image with a feed-forward pipeline. Finally, our framework uses the masks generated from the mask generator to adaptively choose HR patches and recovers them into an HR full image. 

\subsection{Difficulty Identifier Module}

Difficulty identifier module(DIM) is the key component for our framework to reconstruct images, the performance of dual-way SR network lies on the accuracy of DIM. 
As shown in Figure~\ref{fig:example}, we visualize some results of Bicubic interpolation. It is obvious that the hard patch tends to exhibit higher PSNR value as simple/plain patches show lower PSNR. Based on this observation, we use the Bicubic PSNR score as its SR difficulty indicator. However, PSNR is a full-reference assessment metric, thus it is difficult for us to identify SR difficulty by PSNR during test phase. To exploit the PSNR prior, we model the problem of SR difficulty identification as a classification problem and train a difficulty identifier by using LeNet~\cite{lenet} as the basic backbone of DIM. 
Specifically, we adopt Bicubic PSNR value as the target to train the DIM. First, we cropped LR and HR pairs from our training dataset and reconstructed the upscaled patches by Bicubic. Then, we compute the Bicubic PSNR values of all samples and categorize them values into 5 classes, where each class represent a difficulty level. Suppose we translate the difficulty level of a patch to a one-hot vector according to its Bicubic PSNR, and let $\mathbf{y}$ be a one-hot ground-truth label vector, $\mathbf{x}$ be an input patch and $C$ be the set of 5 possible difficulty labels. Then we use a network $F_{\mathbf{W}}$ parameterized by weights $\mathbf{W}$ as our difficulty identifier to learn a function mapping a patch to its difficulty level. Therefore, our goal is to find weights $\mathbf{W}^*$ to minimize the following softmax cross entropy loss function, which is used in our method:
\begin{equation}
\begin{split}
	L(\hat{\mathbf{y}},\mathbf{y},\mathbf{W}) = - \frac{1}{\left | C \right |}\sum_{c\epsilon C}y_{c}log\hat{y_c},
\end{split}
\end{equation}
where
\begin{equation}
\begin{split}
	\hat{\mathbf{y}} = softmax(z) = \frac{exp(z)}{\sum_{c\epsilon C}exp(z_c)},
\end{split}
\end{equation}
and z is non-transformed logit output of our difficulty identifier. 

Let \{$c_1,...,c_{5}$\} represents the 5 difficulty levels, and the difficulty level is ordered by the index. The greater the index, the harder it is. The output of our difficulty identifier for each patch is a probability vector $\hat{\mathbf{y}}$ =\{$p_1,...,p_{5}$\}, where $p_i$ represents the possibility a patch falls in difficulty level $c_i$, $i=1,2,...,5$.

\subsection{Mask Generator}
Our mask generator is used to generate mask by the probability vector from the difficulty identifier to help our framework adaptively enforce dual-way SR to yield superior results. Let $\hat{\mathbf{y}}$ =\{$p_1,...,p_{5}$\} represent the probability output of an LR patch after it is passed into our difficulty identifier, where $p_i$ represents the possibility a patch falls in difficulty level $c_i$, $i=1,2,...,5$. If $p_1$ is the maximum value in vector $\hat{\mathbf{y}}$, the mask generator generates a one mask with the size of its corresponding reconstructed patch, otherwise, the mask generator generates a zero mask. Therefore, our mask generator can be model as:
\begin{equation}
{mask}(\hat{\mathbf{y}}) = \left\{ \begin{array}{l}
1,\; max(\hat{\mathbf{y}})\ ==\ p_1\ ;\\
0,\;otherwise;\\
\end{array} \right.\\
\end{equation}

With the help of the mask generator, we can adaptively enforce the dual-way SR as follow:
\begin{equation}
\begin{split}
	I_{sr} = (1 - mask) \times F_{CB}(I_{lr}) \\
	+ mask \times F_{PB}(I_{lr})
\end{split}
\end{equation}
where $I_{sr}$, $F_{CB(I_{lr})}$, $F_{PB(I_{lr})}$ are the reconstructed patch and $\times$ represents dot multiplication.

\begin{table*}[ht]
\centering
\footnotesize
\caption{The PSNR and SSIM results of different approaches on Set5, Set14, BSDS100 and Urban100 with down-sampling factor $\times$2, $\times$3 and $\times$4. We use {\color{red}{red}} and {\color{blue} blue} to label first and second place, respectively.}
\resizebox{1\linewidth}{20mm}{
\begin{tabular}{|c|c|c|c|c|c|c|c|c|c|c|c|c|c|c|c|c|c|c|c|}
\hline

\multirow{2}{*}{Dataset} & \multirow{2}{*}{Scale} & \multicolumn{2}{c|}{Bicubic} & \multicolumn{2}{c|}{A+} & \multicolumn{2}{c|}{SRCNN~\cite{srcnn}} & \multicolumn{2}{c|}{FSRCNN~\cite{fsrcnn}} & \multicolumn{2}{c|}{VDSR~\cite{vdsr}} & \multicolumn{2}{c|}{LapSRN~\cite{lapsrn}} & \multicolumn{2}{c|}{MemNet~\cite{memNet}} & \multicolumn{2}{c|}{IDN~\cite{idn}} & \multicolumn{2}{c|}{Our} \\ \cline{3-20}
	&	& \multicolumn{1}{c|}{PSNR} & \multicolumn{1}{c|}{SSIM} & \multicolumn{1}{c|}{PSNR} & \multicolumn{1}{c|}{SSIM} & \multicolumn{1}{c|}{PSNR} & \multicolumn{1}{c|}{SSIM} & \multicolumn{1}{c|}{PSNR} & \multicolumn{1}{c|}{SSIM} & \multicolumn{1}{c|}{PSNR} & \multicolumn{1}{c|}{SSIM} & \multicolumn{1}{c|}{PSNR} & \multicolumn{1}{c|}{SSIM} & \multicolumn{1}{c|}{PSNR} & \multicolumn{1}{c|}{SSIM} & \multicolumn{1}{c|}{PSNR} & \multicolumn{1}{c|}{SSIM} & \multicolumn{1}{c|}{PSNR} & \multicolumn{1}{c|}{SSIM} \\\hline
	
\multirow{3}{*}{Set5} & $\times$2 & 33.66 & 0.9299 & 36.54 & 0.9544 & 36.66 & 0.9542 & 37.00 & 0.9558 & 37.53 & 0.9587 & 37.52 & 0.9591 & \textcolor{blue}{37.83} & \textcolor{blue}{0.9600} & 37.78 & 0.9597 & \textcolor{red}{37.87} & \textcolor{red}{0.9600} \\
 & $\times$3 & 30.39 & 0.8682 & 32.58 & 0.9088 & 32.75 & 0.9090 & 33.16 & 0.9140 & 33.66 & 0.9213 & 33.81 & 0.9220 & \textcolor{blue}{34.11} & \textcolor{red}{0.9253} & 34.09 & 0.9248 & \textcolor{red}{34.17} & \textcolor{blue}{0.9252} \\
 & $\times$4 & 28.42 & 0.8104 & 30.28 & 0.8603 & 30.48 & 0.8628 & 30.71 & 0.8657 & 31.35 & 0.8838 & 31.54 & 0.8852 & \textcolor{red}{31.82} & \textcolor{red}{0.8903} & 31.74 & 0.8893 & \textcolor{blue}{31.81} & \textcolor{blue}{0.8890} \\ \hline
 
\multirow{3}{*}{Set14} & $\times$2 & 30.24 & 0.8688 & 32.28 & 0.9056 & 32.42 & 0.9063 & 32.63 & 0.9088 & 33.03 & 0.9124 & 32.99 & 0.9124 & \textcolor{blue}{33.30} & \textcolor{blue}{0.9148} & 33.28 & 0.9142 & \textcolor{red}{33.39} & \textcolor{red}{0.9161} \\
 & $\times$3 & 27.55 & 0.7742 & 29.13 & 0.8188 & 29.28 & 0.8209 & 29.43 & 0.8242 & 29.77 & 0.8314 & 29.79 & 0.8325 & 29.99 & \textcolor{blue}{0.8354} & \textcolor{blue}{30.00} & 0.8350 & \textcolor{red}{30.02} & \textcolor{red}{0.8370} \\
 & $\times$4 & 26.00 & 0.7027 & 27.32 & 0.7491 & 27.49 & 0.7503 & 27.59 & 0.7535 & 28.01 & 0.7674 & 28.09 & 0.7700 & 28.25 & \textcolor{blue}{0.7730} & \textcolor{blue}{28.26} & 0.7723 & \textcolor{red}{28.29} & \textcolor{red}{0.7750} \\ \hline

\multirow{3}{*}{B100} & $\times$2 & 29.56 & 0.8431 & 31.21 & 0.8863 & 31.36 & 0.8879 & - & - & 31.90 & 0.8960 & 31.80 & 0.8952 & \textcolor{blue}{32.08} & \textcolor{blue}{0.8985} & \textcolor{blue}{32.08} & 0.8978 & \textcolor{red}{32.11} & \textcolor{red}{0.8988} \\
 & $\times$3 & 27.21 & 0.7385 & 28.29 & 0.7835 & 28.41 & 0.7863 & - & - & 28.82 & 0.7976 & 28.82 & 0.7980 & 28.95 & \textcolor{blue}{0.8013} & \textcolor{blue}{28.96} & 0.8001 & \textcolor{red}{28.98} & \textcolor{red}{0.8024} \\
 & $\times$4  & 25.96 & 0.6675 & 26.82 & 0.7087 & 26.90 & 0.7101 & - & - & 27.29 & 0.7251 & 27.32 & 0.7275 & \textcolor{blue}{27.41} & \textcolor{blue}{0.7297} & 27.40 & 0.7281 & \textcolor{red}{27.43} & \textcolor{red}{0.7312} \\ \hline
 
\multirow{3}{*}{Urban100} & $\times$2 & 26.88 & 0.8403 & 29.20 & 0.8938 & 29.50 & 0.8946 & - & - & 30.76 & 0.9140 & 30.41 & 0.9103 & 31.27 & \textcolor{blue}{0.9196} & \textcolor{blue}{31.31} & 0.9195 & \textcolor{red}{31.77} & \textcolor{red}{0.9247} \\
 & $\times$3 & 24.46 & 0.7349 & 26.03 & 0.7973 & 26.24 & 0.7989 & - & - & 27.14 & 0.8279 & 27.07 & 0.8275 & 27.42 & 0.8359 & \textcolor{blue}{27.56} & \textcolor{blue}{0.8376} & \textcolor{red}{27.78} & \textcolor{red}{0.8434} \\
 & $\times$4 & 23.14 & 0.6577 & 24.32 & 0.7183 & 24.52 & 0.7221 & - & - & 25.18 & 0.7524 & 25.21 & 0.7562 & 25.41 & \textcolor{blue}{0.7632} & \textcolor{blue}{25.50} & 0.7630 & \textcolor{red}{25.71} & \textcolor{red}{0.7725} \\ \hline
\end{tabular}}
\label{tab:comparison}
\end{table*}

\subsection{Complex Branch and Plain Branch}
In our approach, CB is used to recover hard images while we use PB for reconstructing easy images. We choose IDN~\cite{idn} as the backbone of our CB as IDN is an efficient SR framework with competitive performance. Since Bicubic interpolation demonstrates superior performance in plain area with high efficiency, we adopt Bicubic interpolation as our PB. There are some advantages to our multi-branched SR framework. First, patch-wise SR can take full advantage of the powerful parallelism computation of GPU by reconstructing high-resolution patches in batch. Second, we can embrace the SR ability of different models for more accurate super-resolution. For example, we can use a heavy model, such as EDSR, as the backbone of CB to reconstruct hard image patches more accurately while using a light-weight model as the backbone of PB to processing easy/plain image patches.

\section{Experiments}
\label{sec:exper}
We first clarify the experimental settings about datasets, degradation models, evaluation metric, and training settings.

\subsection{Datasets and Evaluation metrics}
Following ~\cite{ntire2017}, we use DIV2K as training set. For testing, we use four standard benchmark datasets, i.e., Set5, Set14, BSD100, Urban100~\cite{urban100} for evaluation. We obtain the LR input with Bicubic downsample. Besides, We conduct the evaluation with PSNR and SSIM metrics on Y channel (i.e., luminance) of transformed YCbCr space.

\subsection{Implementation details}
Our framework adopts LeNet-5~\cite{lenet} as the network backbone of our difficulty identifier, IDN~\cite{idn} as the  network backbone of our CB and Bicubic interpolation as our PB. The training procedure of our framework can be divided into two parts. The first part is the end-to-end training of our DIM. We use the dataset based on PSNR prior built by Bicubic to train the DIM. Another part is jointly end-to-end training the CB and PB with the help of DIM and mask generator. All patches will be passed through the DIM, CB, and PB to generate corresponding results. Only the reconstructed patches with one mask will be used to compute the loss and gradients of back-propagation, then the parameter of the corresponding SR branch is updated with gradients. 

Data augmentation is performed on DIV2K training set, which are randomly rotated by $90^\circ$, $180^\circ$, $270^\circ$ and flipped horizontally. In each training batch, 64 LR patches with the size of 48$\times$48 are extracted as inputs. Our model is trained by ADAM optimizer~\cite{adam} with  $\beta_1$ = 0.9, $\beta_2$ =0.999, and $\epsilon$  = $10^{-8}$. The initial learning rate is set to 1$e-$4 and then decreases to half every 100 epoch. We use PyTorch~\cite{pytorch} to implement our models with a Titan XP GPU.

\begin{figure*}[t] 
    \centering
	\centerline{\includegraphics[width=1\linewidth]{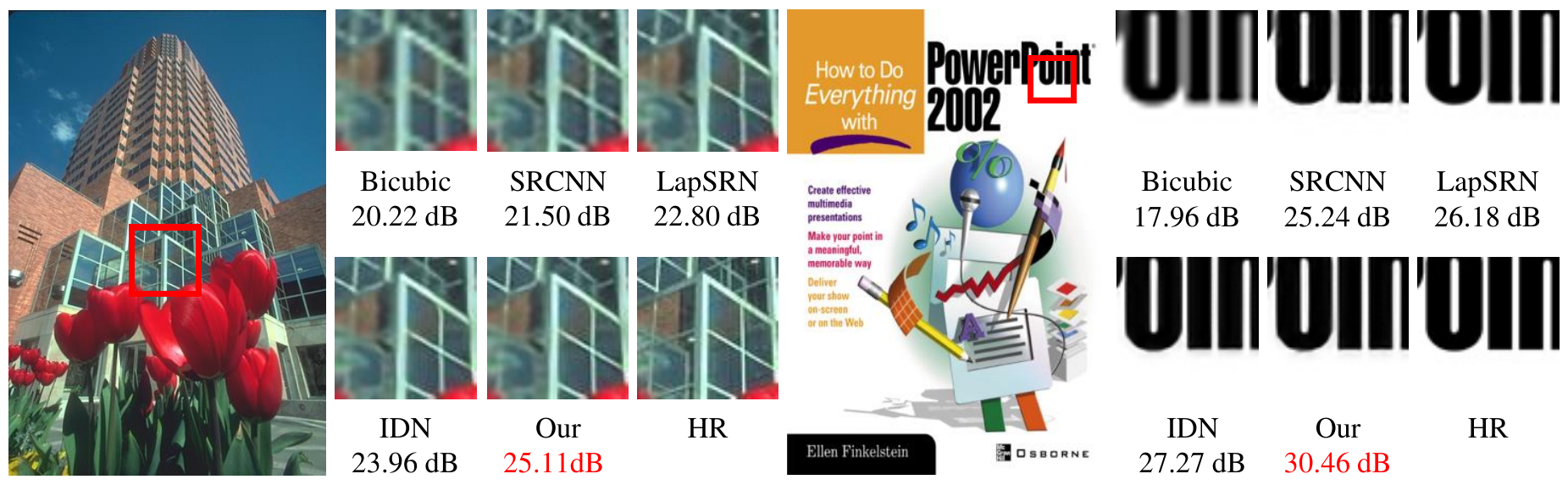}}
	\caption{Qualitative comparisons on ``89000'' image from BSD100 and ``ppt3'' from Set14.}
	\label{fig:visual}
	\vspace{-1mm}
\end{figure*}
\subsection{Comparison with state-of-the-art methods}
We compare our approach with state-of-the-art SR methods mainly on two commonly-used image quality metrics: PSNR and the structural similarity index(SSIM). The state-of-the-art methods we used are Bicubic, A+, SRCNN~\cite{srcnn}, FSRCNN~\cite{fsrcnn}, VDSR~\cite{vdsr}, LapSRN~\cite{lapsrn}, MemNet~\cite{memNet}, and IDN~\cite{idn}. 

Table~\ref{tab:comparison} shows the quantitative comparisons on the Set5, Set14, B100, and Urban100 with factor $\times$2, $\times$3 and $\times$4. As illustrated in Table~\ref{tab:comparison}, our approach surpasses IDN with a clear margin. Specifically, our approach outperforms IDN with 0.11 dB and 0.46 dB under factor $\times$2 on Set14 and Urban100, respectively. Moreover, our full model surpasses IDN with 0.22 dB and 0.08 dB on Set5 and Urban100 under factor $\times$3. As IDN is an efficient SR framework, our model still owns a superiority among state-of-the-art methods, which inherently verify the effectiveness of our model. When compared with the remaining models, our approach outperforms them by a large margin on Urban100 in term of the PSNR metric. A similar trend can also be observed for the SSIM score. Specifically, as illustrated in Table~\ref{tab:comparison}, our approach achieves 0.50 dB, 0.36 dB, and 0.30 dB improvement over MemNet~\cite{memNet} on Urban100.

Figure~\ref{fig:visual} visualizes some promising examples from B100 and Set14. We interpolate the Cb and Cr chrominance channels by the Bicubic method and translate YCbCr space to RGB space to generate color images for better views. We can observe that our approach restores sharper and clearer images with higher PSNR than other methods. As shown in Figure~\ref{fig:visual}, our model restore a clear structure with fewer artifacts. 

\subsection{Efficiency}
We demonstrate an efficiency comparison to verify the practicability of our framework. As shown in Table~\ref{tab:efficiency}, we conduct this efficiency analysis on Urban100 with factor $\times$4. Since DIM needs additional computational cost, our full model obtains a parameter gain and slower than IDN. However, our model still demonstrates a competitive efficiency among state-of-the-art image SR methods. For instance, compared LapSRN, our model achieves faster efficiency with fewer parameters. 

\begin{table}[t]
\centering
\footnotesize
\caption{Efficiency comparison on Urban100 with factor $\times$4. }
\begin{tabular}{|c|c|c|c|c|}
\hline
Algorithm      & VDSR  & LapSRN & IDN   & Our   \\ \hline
Time(s/frame)  & 0.094 & 0.046  & 0.015 & 0.031 \\ \hline
Parameter(MB)  & 2.824  & 3.327  & 2.597 & 3.226 \\ \hline
\end{tabular}
\label{tab:efficiency}
\vspace{-1mm}
\end{table}

\subsection{Ablation study of PB and CB}
\begin{table}[t]
\centering
\footnotesize
\caption{Investigations of PB and CB on Urban100. Best results are highlighted with \textbf{boldface}. We can observe that ``PB+CB'' combination achieve the best performance. }
\begin{tabular}{|c|c|c|c|c|}
\hline
\multirow{2}{*}{Dataset} & \multirow{2}{*}{Scale} & PB & CB & PB + CB \\ \cline{3-5}
	&	& PSNR & PSNR & PSNR \\\hline
\multirow{3}{*}{Urban100} & $\times$2 & 29.80 & 34.42 & \textbf{34.62}  \\
 & $\times$3 & 27.59 & 30.49 & \textbf{30.95}  \\
 & $\times$4 & 25.04 & 27.62 & \textbf{27.64}  \\ \hline
\end{tabular}
\label{tab:comparison1}
\vspace{-3mm}
\end{table}

In this section, we study the effects of each branch in our proposed approach, we disable one branch each time and compare their differences on different testing sets. 

\textbf{Effects on the dataset with diverse difficulty}. We first compare the performance of different SR branch on Urban100 benchmark dataset. We crop the LR images of Urban100 into patches with the size of 48$\times$48 and the corresponding HR patches with the size of (48$\times$scale)$\times$(48$\times$scale). We use all the patches as the input of PB, CB, and our integrated adaptive approach(PB + CB). Then we compute the PSNR values between the reconstructed patches and their HR patches. The results are shown in Table~\ref{tab:comparison1}. We can observe that PB and CB have different SR performance on the same testing set. CB can handle better than PB on datasets with diverse difficulty. Although the CB can achieve high PSNR, our dual-way adaptive approach achieves the best performance compared with CB and PB. This shows that our approach is an effective SR approach while handling dataset with diverse difficulty.

\begin{table}[t]
\centering
\footnotesize
\caption{Study of PB and CB on hard/easy patches in Urban100. Best results are highlighted with \textbf{boldface}.}
\begin{tabular}{|c|c|c|c|}
\hline

\multirow{2}{*}{Dataset} & \multirow{2}{*}{Scale} & PB & CB \\ \cline{3-4}
	&	& PSNR & PSNR  \\\hline
\multirow{3}{*}{Hard Patches} & $\times$2 & 28.10 & \textbf{33.06}   \\
 & $\times$3 & 25.70 & \textbf{29.24}   \\
 & $\times$4 & 24.19 & \textbf{26.92}   \\ \hline
 
 \multirow{3}{*}{Easy Patches} & $\times$2 & \textbf{53.95} & 52.71   \\
 & $\times$3 & \textbf{63.66} & 54.25   \\
 & $\times$4 & \textbf{54.26} & 51.86 \\ \hline
\end{tabular}
\label{tab:comparison2}
\vspace{-3mm}
\end{table}
\textbf{Effects on the dataset with single difficulty}. We further show the effect of PB and CB on the dataset with single difficulty. For simplicity, we crop pair patches from Urban100, compute their PSNR values and divide them into two set(hard and easy) by their PSNR values. We set the threshold of PSNR value as 45. If the PSNR value of a pair patch exceeds 45, we divide it into easy patches set, otherwise dividing it into hard patches set. Table~\ref{tab:comparison2} shows the performance of PB and CB on hard/easy patches of Urban100. We can observe that PB and CB have different performance on the dataset with different difficulty. PB handle plain/easy images better than CB while CB process hard images better. This shows that a single model is hard to handle all regions with diverse difficulty well at the same time. We should embrace the different SR ability of different methods to produce more accurate results.

\section{Conclusion}
\label{sec:con}
In this paper, we proposed a novel dual-way adaptive SR approach that can discriminately process each image region of an image by its difficulty.  Extensive experiments on several standard benchmarks demonstrate the effectiveness of our approach. 

\bibliographystyle{IEEEbib}
\bibliography{egbib}

\end{document}